\documentclass[10pt,twocolumn,letterpaper]{article}

\usepackage{wacv}              

\usepackage[accsupp]{axessibility}  
\definecolor{wacvblue}{rgb}{0.21,0.49,0.74}
\usepackage[pagebackref,breaklinks,colorlinks,allcolors=wacvblue]{hyperref}
\usepackage{multirow}

\def\wacvPaperID{1350} 
\def\confName{WACV}
\def\confYear{2027}

\title{ADC-RANet: Improving Resolution Adaptive Neural Networks with \\ Attention, Distillation and Calibration}

\author{
Jeferson Severino de Araujo, Lilian Hollard,
 Eric Granger, Marco Pedersoli\\[0.4em]
LIVIA, Department of Systems Engineering, ETS Montreal, Canada\\
{\tt\small \{jeferson.severino-de-araujo.1\}@ens.etsmtl.ca}\\
{\tt\small \{lilian.hollard, eric.granger, marco.pedersoli\}@etsmtl.ca}
}

\begin{document}
\maketitle
\begin{abstract}

Dynamic neural networks are a class of adaptive models in which the computation assigned to each input sample is determined dynamically, potentially reducing computational cost while maintaining similar performance. For image classification, 
the Resolution Adaptive Network (RANet) is a strong approach that progressively allocates computational resources across input scales based on sample difficulty. In this work, we introduce RANet with Attention, Distillation and Calibration (ADC-RANet), an enhanced version of RANet that improves performance across all computational budgets. Our approach combines three tightly connected components: bottom-up channel attention, top-down cascaded knowledge distillation, and overconfidence correction using MixUp, in a synergistic design that coordinates both pathways. Experiments on CIFAR-100 and ImageNet datasets show consistent gains while preserving the efficiency of RANet. Our code is available: \url{https://github.com/jefersonseverino/ADC-RANet}

\end{abstract}
    
\section{Introduction} \label{sec:intro}

Deep learning has revolutionized computer vision, greatly improving tasks such as object detection and image classification \cite{alexnet, objectdetection}. Meanwhile, connected sensors, mobile platforms, and intelligent edge systems have created a need to deploy these models beyond traditional high-performance computing environments \cite{survey_ahmad,heydari2025tiny}. Many modern applications require real-time, on-device processing, making local inference essential \cite{heydari2025tiny, qin2024mobilenetv4}.


However, the growing demand for deeper, more accurate models is outpacing the computational capabilities of embedded hardware \cite{luoembedded}, making deployment on resource-constrained platforms increasingly challenging. Strict limits on computation, memory, energy, and latency impose strong constraints, especially for real-time vision applications. A key research challenge, therefore, is designing deployment strategies that balance accuracy and efficiency, enabling low-power systems to meet hardware constraints while maintaining competitive performance.

In this context, dynamic neural networks have emerged as a cost-effective alternative to conventional static architectures \cite{han2022dynamic}. Unlike traditional networks, which incur a fixed computational cost per input, dynamic models adapt their computation to sample difficulty. They rely on multiple auxiliary classifiers, or multi-exit designs, at different depths, allowing them to stop inference early once a confidence threshold is reached and thereby reducing the overall computation.

 \begin{figure}[t]
  \centering
   \includegraphics[width=1.0\linewidth]{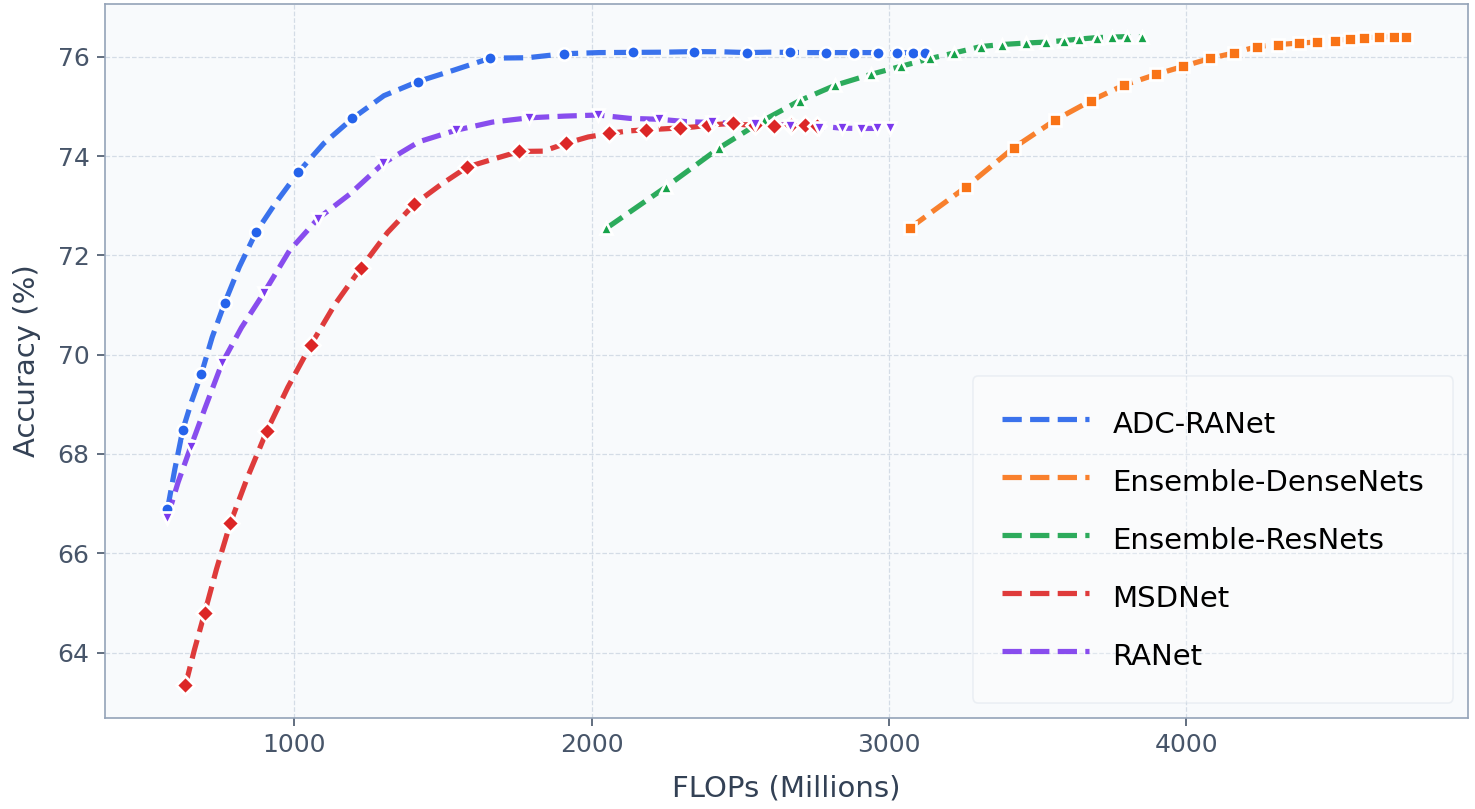}
    
   \caption{
   Budgeted batch prediction on ImageNet, reporting accuracy at different compute budgets. Our method, ADC-RANet, consistently outperforms MSDNet \cite{huang2018multiscale} and RANet \cite{yang2020resolution} at comparable budgets, while nearing ResNet \cite{resnet} and DenseNet \cite{densenet} ensemble performance at much lower computational cost.
   }
   \label{fig:imagenet-bb}
\end{figure}

Most work on multi-exit networks relies on a multi-resolution strategy, where the same input is processed at several resolutions, each paired with its own classifier. Examples include Multi-Scale Dense Network (MSDNet) \cite{huang2018multiscale} and Resolution Adaptive Network (RANet) \cite{yang2020resolution}. 
In this work, we focus on RANet, a dynamic, spatially adaptive network that starts inference at low resolution and progressively increases the resolution only when prediction confidence drops below a threshold. By avoiding unnecessary high-resolution processing, this adaptive strategy is particularly well suited to resource-constrained settings. 

RANet progressively builds predictions by reusing representations across multiple resolutions and refining them through successive classifiers, using prediction confidence to enable early termination whenever possible. Improving this process therefore requires a coordinated approach that addresses the three key components of the architecture: reliable confidence estimation to avoid overconfident early predictions, effective knowledge transfer from later, more accurate classifiers to earlier subnetworks, and improved information flow to strengthen the representations used by later classifiers.

Specifically, we adapt the Channel-Attention Mechanism from \cite{addad2025chase} by adding a residual sum at the end of the block, which we refer to as the Residual Channel Attention Mechanism (RCAM), and integrate it into RANet's fusion blocks to enhance cross-scale feature reuse at minimal computational cost. Among attention-based methods, channel attention is the only approach that remains computationally lightweight enough for low-computation devices \cite{hu2018squeeze}, since it operates in $\mathcal{O}(C^2/r)$ instead of $\mathcal{O}(n^2)$, where $C$ is the number of channels and $n$ is the spatial resolution.



In addition, we introduce a Cascaded Knowledge Distillation (CKD) scheme in which each classifier serves as the teacher for its immediate predecessor. 
Inspired by \cite{mirzadeh2020improved}, which studied how the distillation gap affects distillation in two separate networks, we design classifier (n+1) to be the teacher for classifier (n).
Neighboring classifiers operate at closely related resolutions and produce more similar feature representations; we evaluate whether this strategy works in the shared backbone of RANet.
%
%
%
Furthermore, we incorporate MixUp \cite{zhang2018mixup} during training, aiming to both reduce the overconfidence present in RANet and serve as an augmentation technique to improve the performance of all exits. Finally, we discuss the gradient conflict present in RANet and how the training strategy is important depending on the final deployment objective scenario.


The main contributions of this work are the following: 
\begin{enumerate}

    \item Propose a novel Cascaded Knowledge Distillation loss for RANet that improves early classifiers through a predecessor-supervised hierarchy, and empirically compare it with alternative teacher topologies.
    
    \item Introduce a lightweight bottom-up channel-attention mechanism (RCAM) and a GELU activation replacement to improve late-exit accuracy, along with a MixUp-based strategy that mitigates RANet's overconfidence and further improves late-exit accuracy.
    
    \item Analyze the gradient conflicts between low- and high-resolution classifiers in RANet and show how different training strategies can be selected to prioritize early-exit performance, final-exit accuracy, or a balance between the two, depending on the target deployment scenario.

\end{enumerate}

\section{Related Work}
\label{sec:formatting}



\textbf{Dynamic networks} with multiple exits have emerged as a widely employed approach to adapt computational resources based on input data difficulty, achieving improved computation-performance trade-offs. Multi-exit networks, such as BranchyNet \cite{teerapittayanon2016branchynet}, MSDNet \cite{huang2018multiscale}, and RANet \cite{yang2020resolution}, consist of multiple classifiers positioned at different depths, allowing inference to terminate early whenever the confidence and accuracy of an intermediate prediction are sufficiently high. 
Building on this, \citet{huang2018multiscale} propose MSDNet, a DenseNet-based multiscale architecture that maintains and processes multiple input resolutions in parallel to extract features from different resolutions. 
Yang et al. \cite{yang2020resolution} introduce RANet, a dynamic spatial-adaptive network that starts inference at a low input resolution and scales to higher resolutions only when necessary. The image is processed at progressively higher resolutions: if a low-resolution classifier reaches a confidence threshold, inference stops; otherwise, its intermediate features and the higher-resolution input are passed to the next subnetwork, up to the final, highest-resolution stage.
This strategy further reduces computation: early exits also operate at lower resolutions, lowering cost and making RANet suitable for environments with strict resource constraints.

Several works improve multi-exit performance via \textbf{architectural changes}. \citet{addad2023multiexit} enhance MSDNet by redesigning feature fusion blocks to strengthen cross-scale information flow and feature reuse. \citet{addad2025chase} extend this by adding global attention to MSDNet. While these methods boost feature interactions, they target multiscale architectures that process all scales simultaneously, and their additional overhead can conflict with the core goal of multi-exit networks: reducing inference costs. In contrast, our work focuses on resolution-adaptive architectures, specifically RANet, and studies lightweight channel-attention mechanisms that enable cross-scale feature reuse with minimal computational overhead.


\noindent \textbf{Adaptive spatial processing for dynamic inference.} 
\citet{wang2020glance} propose a two-stage glance-and-focus framework that makes a low-resolution prediction and, when confidence is insufficient, uses reinforcement learning to locate discriminative regions for refinement, at the cost of added training complexity. \citet{zhu2021dynamic} instead learn a resolution predictor that selects input resolution for transformers without hurting performance. Neither method captures collaboration between classifiers across resolutions.

Beyond architectural improvements, several studies have focused on \textbf{training strategies} for multi-exit networks. \citet{han2022learning} proposes a meta-learning framework that dynamically adjusts the importance of each exit during training. Similarly, \citet{gong2024deep} studies gradient conflicts in MSDNet and proposes a deep-feature optimization strategy to improve joint learning across exits. These approaches improve optimization, but do not explicitly enhance knowledge transfer between exits.

\noindent \textbf{Knowledge distillation} is a dominant approach for improving multi-exit networks. Early work by \citet{phuong2019distillation} trains shallow exits to mimic the final classifier, while \citet{wang2021harmonized} proposes Harmonized Dense Knowledge Distillation (HDKD), where each exit learns from all subsequent exits. \citet{zhang2024multilevel} aggregates information from all classifiers through a weighted framework. \citet{guidez2025erde} proposes an entropy-based distillation strategy for multi-exit architectures, but this approach requires separately trained networks. Despite their benefits, these methods usually rely on a strong final classifier as the primary teacher and {ignore the complexity distance between classifiers}. This overlooks results by \citet{mirzadeh2020improved}, who show that a large teacher–student gap can harm knowledge transfer, potentially imposing overly difficult learning targets on early classifiers.



\noindent \textbf{Confidence calibration} is a key challenge for multi-exit architectures because early-exit decisions depend directly on prediction confidence. \citet{guo2017calibration} shows that deep networks tend to be overconfident, and \citet{addad2025seednet} highlights the negative impact of poor calibration on early-exit decisions. Yet calibration is often treated separately from architectural design and knowledge distillation strategies.


Overall, existing approaches primarily address architectural modifications, optimization strategies, knowledge distillation, or calibration as separate aspects of multi-exit networks. To the best of our knowledge, no prior work has systematically investigated these complementary challenges within a resolution-adaptive architecture such as RANet. In this work, we jointly address: (i) enabling efficient cross-resolution feature interaction between subnetworks; (ii) reducing the teacher–student gap through cascaded knowledge distillation across teacher topologies; (iii) mitigating overconfidence to improve the reliability of exit decisions; 
and (iv) evaluating training strategies under different deployment scenarios. Our work addresses these complementary aspects while preserving the computational efficiency that originally motivated multi-exit architectures.

 \section{Proposed ADC-RANet Method}



This section first provides a brief overview of RANet, the baseline architecture used in this work, and then describes the architectural modifications and training enhancements introduced in the proposed ADC-RANet model. 
Each component introduced in this section is motivated by a specific limitation of RANet identified through empirical analysis. CKD is motivated by the gradient conflict that reduces early-exit performance and is designed to reduce computational cost; MixUp mitigates RANet's overconfidence; RCAM is introduced to enhance feature reuse, and GELU is adopted to alleviate the high prevalence of dead neurons observed in the ConvUp operations.


\subsection{Resolution Adaptive Network}

RANet is a multi-scale architecture that enables feature reuse across subnetworks operating at different resolutions. In the cross-scale interaction, a higher-resolution subnetwork leverages intermediate feature maps from lower-resolution subnetworks \cite{yang2020resolution}. More precisely, each dense block receives intermediate representations from the previous block, allowing features extracted at one scale to be upsampled and fused into the next scale.

This fusion process concatenates the feature maps of the current subnetwork with those from the previous one: 
\begin{equation}
\label{eq3}
\mathbf{x}^{s,h}_{\text{out}} =
\mathcal{T}\!\left(
    \mathbf{x}^{s,h}_L \oplus
    \mathcal{F}_{\uparrow}\!\left(
        \mathcal{R}\!\left(\mathbf{x}^{s-1,h-1}_{L-1}\right)
    \right)
\right).
\end{equation}
where $\mathbf{x}^{s,h}_L$ denotes the features of the current dense block before fusion, and $\mathbf{x}^{s-1,h-1}_{L-1}$ corresponds to the features produced at the previous resolution. A pointwise convolution, denoted by $\mathcal{R}$, is applied to reduce the feature map to a lower channel dimensionality and match the input requirements of the subsequent stage. When the previous subnetwork operates at a lower spatial resolution, an upsampling operator $\mathcal{F}_{\uparrow}$ is applied before concatenation. The operator $\oplus$ denotes the channel-wise concatenation and $\mathcal{T}$ represents the DenseNet transition layer.


\subsection{Residual Channel Attention Mechanism}


Self-attention \cite{vaswani2017attention} is state-of-the-art in many deep learning subfields \cite{dosovitskiy2021image}. However, in vision, such mechanisms are computationally costly, with quadratic complexity $\mathcal{O}(n^2)$, where $n$ is the number of input patches. This makes their use on resource-constrained devices impractical. To mitigate this, we adopt a lightweight channel-attention mechanism based on Squeeze-and-Excitation (SE) \cite{hu2018squeeze}. Unlike self-attention, the SE blocks model channel-wise dependencies with complexity $\mathcal{O}(C^2/r)$, where $C$ is the number of channels and $r$ is the channel reduction rate, which controls the bottleneck width in the RCAM block.

Dai et al. \cite{dai2021attentional} show that simple concatenation is suboptimal for fusing features across different spatial scales. In RANet, feature fusion similarly relies on concatenation followed by channel compression, treating low- and high-resolution features uniformly and therefore lacking explicit context awareness. To address this limitation, we introduce RCAM as a context-aware fusion module that operates on the concatenated low- and high-resolution features.



The RCAM operates in two stages. First, global average pooling is applied to each channel to produce a channel descriptor $z_c$. Next, this descriptor is passed through a small gating network, where $z$ denotes the pooled channel vector, $\delta$ is the GELU activation function, and $\sigma$ is the sigmoid activation function:

\begin{equation}
    z_c = \frac{1}{HW}\sum_{i,j} X_c(i,j), \qquad
    \mathbf{s} = \sigma\!\left(\mathbf{W}_2\,\delta(\mathbf{W}_1\mathbf{z})\right),
    \label{eq1}
\end{equation}
where $\mathbf{z}=[z_1,\ldots,z_C]^\top$ is the vector of channel descriptors.

Second, the resulting channel weights $\mathbf{s}$ are used to rescale the input feature map, yielding the refined output $\tilde{X}$.  The residual connection ensures every channel retains its original signal, while the learned gate $\mathbf{s_c}$ determines the relative degree of amplification each channel receives.
\begin{equation}
\label{eq2}
    \tilde{\mathbf{X}}
    =
    \mathbf{s} \odot \mathbf{X}
    +
    \mathbf{X}.
\end{equation}

By applying RCAM before channel compression, the model can emphasize the most informative channels while preserving the computational efficiency required by RANet. The final fused representation is then processed by the compression operator: 
\begin{equation}
\label{eq4}
\mathbf{x}^{s,h}_{\text{out}} =
\mathcal{T}\!\left(
    \operatorname{RCAM}\!\left(
        \mathbf{x}^{s,h}_L \oplus
        \mathcal{F}_{\uparrow}\!\left(
            \mathcal{R}\!\left(
                \mathbf{x}^{s-1,h-1}_{L-1}
            \right)
        \right)
    \right)
\right).
\end{equation}
%
RCAM increases the overall computational cost only by 0.30\% at the highest budget and as low as 0.1\% for early exit on ImageNet, representing a marginal overhead relative to the total computational budget.

\subsection{Cascaded Knowledge Distillation}

Knowledge distillation \cite{hinton2015distilling} is a technique in which a large model, the teacher, supervises a smaller student model. In this work, the student is trained using the teacher’s logits, allowing it to learn from the teacher’s soft output representations rather than hard labels.  
In RANet, we observed a gradient conflict caused by the misalignment (low cosine similarity) between the gradients of high- and low-resolution classifiers (more details in supplementary material). Furthermore, RANet's logic follows a bottom-up approach, where subsequent exits leverage features generated by lower subnetworks. This motivates a complementary top-down strategy, in which later classifiers transfer their knowledge to earlier exits through their predictions to help them improve their performance.




Therefore, we propose CKD, in which each classifier $k$ serves as a teacher for the preceding classifier $k-1$. This neighboring distillation strategy is motivated by the relatively small discrepancy between teacher and student in both spatial resolution and feature representation, which facilitates knowledge transfer between adjacent classifiers.
%
%
%
Each model's logit is defined by $z_k(x)$. The softmax function is applied to the logits to obtain a probability distribution over classes:
\begin{align}
p_k^{T}(x) &= \mathrm{softmax}\!\left(\frac{z_k(x)}{T}\right).
\end{align}
In our case, we compute the softmax function with $T = 1$ for the cross-entropy loss and the temperature-scaled softmax to obtain soft targets for the knowledge distillation loss.


For each exit $k \in {1, \dots, N-1}$, the loss is defined as the sum of a cross-entropy term and a knowledge distillation term. The distillation component follows the KL divergence formulation from \cite{hinton2015distilling}, computed between the temperature-scaled softmax outputs of two consecutive exits, $k$ and $k+1$. It is weighted by a factor $\beta T^2$, where $\beta$ and $T$ are hyperparameters controlling the strength of the distillation. For each exit $k \in \{1,\ldots,N-1\}$, the loss is defined as: 
\begin{equation}
\label{eq6}
    L_k =
    L_{\mathrm{CE}}(p_k,y)
    + \frac{\beta T^2}{N-1}
    \mathrm{KL}\!\left(
        p_{k+1}^{T} \,\|\, p_k^{T}
    \right).
\end{equation} 
We detach the teacher logits when computing the KL loss, and the final classifier is not involved in any distillation process; it is trained solely using the cross-entropy loss.
\begin{equation}
\label{eq7}
    L_N = L_{\mathrm{CE}}(p_N, y).
\end{equation}
%
%
The total loss function of ADC-RANet is defined by:
\begin{equation}
\label{eq8}
L(x,y)
=
L_N
+
\sum_{k=1}^{N-1} L_k
\end{equation}


\subsection{MixUp-Based Calibration}


Calibration is essential for deploying multi-exit networks on resource-constrained devices \cite{guo2017calibration}, as it measures how well a network’s confidence aligns with its accuracy. Overconfident classifiers can trigger premature exits by assigning high confidence to incorrect predictions, reducing overall accuracy. As shown in Table \ref{tab:signed_bias_comparison}, RANet is systematically overconfident across its exits. We quantify this using the signed calibration bias for exit $k$, defined as $\text{SB}_k = \text{conf}(k) - \text{acc}(k)$, where $\text{conf}(k)$ is the mean post-softmax confidence and $\text{acc}(k)$ is the corresponding accuracy. A positive $\text{SB}_k$ indicates overconfidence, and a negative value indicates underconfidence.


To improve calibration in our setting, we adopt MixUp \cite{zhang2018mixup}. It generates augmented samples by interpolating both inputs and labels. Each interpolated sample is downsampled to the resolutions required by the different subnetworks, allowing the same augmented sample to be used across all exits. This approach provides the calibration benefits of MixUp while simultaneously serving as a data augmentation strategy, without introducing additional inference-time computation or requiring post-hoc calibration. We name it MBC (MixUp-Based Calibration).

\subsection{Method Overview}
\label{overview}

ADC-RANet integrates three components that jointly improve the efficiency–accuracy trade-off of the baseline RANet. At the feature level, RCAM enhances the bottom-up pathway by emphasizing informative channels before compression, boosting cross-scale feature reuse with minimal overhead. At the supervision level, CKD adds a top-down pathway where each classifier guides its predecessor, reducing gaps between exits and progressively transferring knowledge along the subnetwork. In addition, by a simple change in the activation function from ReLU to GELU, we improve feature reuse, which in turn enhances exit performance. 
%
%
ADC-RANet adds three complementary components to the baseline RANet that work in synergy, improving its efficiency-accuracy trade-off across computational budgets without sacrificing its lightweight design. We call ADC-RANet the model with the three components (RCAM, CKD, and MBC) activated with the GELU activation function. 
\section{Results and Discussion}

This section presents a comparative evaluation of ADC-RANet and early-exit networks on CIFAR-100 and ImageNet, considering both anytime prediction and budgeted batch inference settings. We further conduct an ablation study of the proposed components to assess their individual contributions and analyze their interactions. Finally, we investigate the impact of gradient conflicts across the different subnetworks of RANet and examine how the proposed CKD strategy mitigates these effects.

\subsection{Experimental Setup}
\label{sec:exp_setup}


\noindent \textbf{Datasets.} CIFAR-100 is a 32 x 32 RGB dataset with 100 classes. It contains 50,000 images for training and 10,000 for validation.  ImageNet is a 224 x 224 RGB dataset from ILSVRC2012 \cite{deng2009imagenet} with 1000 classes. It contains about 1.2 million training images and 50,000 for validation.

\noindent \textbf{Metrics.} Our models are evaluated using anytime prediction and budgeted batch prediction metrics \cite{huang2018multiscale}. The anytime metric characterizes the accuracy achievable when stopping at a given exit K. 
It measures the performance of each exit independently over the full test set. The budgeted batch metric reflects performance under a fixed average computational cost. 
For budgeted batch prediction, thresholds are calibrated on a subset of the training data and then assessed on the validation set, following \cite{yang2020resolution}. 

\noindent \textbf{Hyperparameters.}
We follow the original RANet training setup (hyperparameters, policy, and data augmentation) \cite{yang2020resolution} for 300 epochs across all experiments and baselines on CIFAR-100 and 90 epochs on ImageNet, with a batch size of 64 for CIFAR-100 and 512 for ImageNet, and a multistep learning schedule that reduces the learning rate by a factor of 10 at epochs 150 and 225 for CIFAR-100 and every 30 epochs for ImageNet. We have three modifications from \cite{yang2020resolution}. First, we add MixUp with parameter $\alpha=2$, so each sample is a convex combination of two images. Second, for RCAM we set the reduction rate to $r=2$, and for CKD we use $\beta=5$ with temperature $T=5$. MSDNet follows the same training as \cite{huang2018multiscale}. The ResNet ensemble baseline was trained for 300 epochs, and the DenseNet ensemble baseline for 300 epochs.


\begin{table}[!t]
\centering
\caption{Signed Bias across exits for MixUp, Temperature Scaling, and RANet for three seeds.}
\label{tab:signed_bias_comparison}
\small
\setlength{\tabcolsep}{5pt}
\begin{tabular}{c|ccc}
\hline
\textbf{Exit} & \textbf{MixUp} & \textbf{Temp.} & \textbf{RANet} \\
\hline
@1 & $+0.01\!\pm\!0.01$ & $+0.07\!\pm\!0.01$ & $+0.07\!\pm\!0.01$ \\
@2 & $+0.10\!\pm\!0.05$ & $+0.25\!\pm\!0.02$ & $+0.29\!\pm\!0.02$ \\
@3 & $+0.16\!\pm\!0.04$ & $+0.33\!\pm\!0.03$ & $+0.37\!\pm\!0.03$ \\
@4 & $+0.22\!\pm\!0.04$ & $+0.39\!\pm\!0.01$ & $+0.46\!\pm\!0.01$ \\
@5 & $+0.21\!\pm\!0.04$ & $+0.40\!\pm\!0.02$ & $+0.48\!\pm\!0.02$ \\
@6 & $+0.18\!\pm\!0.05$ & $+0.37\!\pm\!0.03$ & $+0.45\!\pm\!0.03$ \\
@7 & $+0.15\!\pm\!0.08$ & $+0.39\!\pm\!0.06$ & $+0.47\!\pm\!0.02$ \\
@8 & $+0.11\!\pm\!0.07$ & $+0.28\!\pm\!0.01$ & $+0.40\!\pm\!0.04$ \\
\hline
\end{tabular}
\end{table}
\begin{figure}[t]
  \centering
   \includegraphics[width=1.0\linewidth]{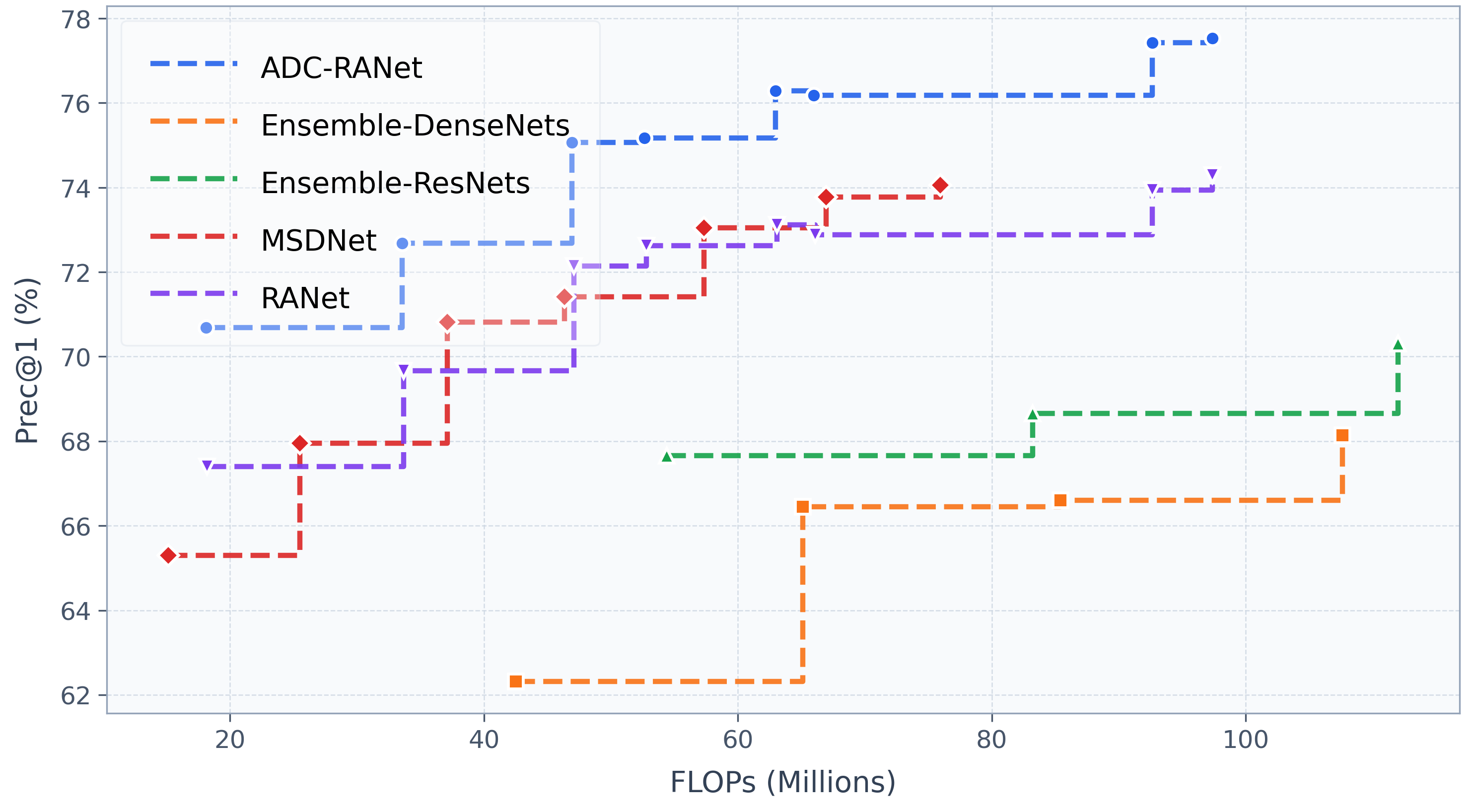}
    
   \caption{Anytime prediction CIFAR-100, showing the accuracy of the full test set at each exit.}
   \label{fig:cifar100anytime}
\end{figure}

\begin{table*}[t!]
\centering
\scriptsize
\setlength{\tabcolsep}{4pt}
\caption{Ablation anytime prediction Top-1 accuracy (\%) on CIFAR-100 over three different seeds.}
\label{tab:anytime_acc}
\begin{tabular}{lcccccccc}
\toprule
Configuration & Exit 1 & Exit 2 & Exit 3 & Exit 4 & Exit 5 & Exit 6 & Exit 7 & Exit 8 \\
\midrule
Baseline (RANet) & $67.40 \pm 0.31$ & $69.66 \pm 0.37$ & $72.14 \pm 0.27$ & $72.62 \pm 0.36$ & $73.12 \pm 0.44$ & $72.88 \pm 0.18$ & $73.94 \pm 0.42$ & $74.29 \pm 0.13$ \\

GELU & $67.33 \pm 0.42$ & $69.82 \pm 0.34$ & $72.70 \pm 0.30$ & $73.14 \pm 0.67$ & $73.82 \pm 0.49$ & $73.84 \pm 0.25$ & $74.76 \pm 0.30$ & $75.41 \pm 0.11$ \\
RCAM & $67.66 \pm 0.03$ & $70.31 \pm 0.46$ & $72.56 \pm 0.23$ & $72.73 \pm 0.06$ & $73.68 \pm 0.32$ & $73.82 \pm 0.16$ & $75.25 \pm 0.17$ & $75.69 \pm 0.10$ \\
MBC & $68.40 \pm 0.16$ & $71.12 \pm 0.21$ & $73.65 \pm 0.18$ & $73.88 \pm 0.04$ & $74.66 \pm 0.26$ & $74.94 \pm 0.32$ & $75.86 \pm 0.15$ & $76.25 \pm 0.11$ \\
CKD & $70.34 \pm 0.28$ & $72.07 \pm 0.02$ & $73.48 \pm 0.12$ & $73.56 \pm 0.13$ & $74.11 \pm 0.08$ & $74.25 \pm 0.16$ & $74.62 \pm 0.14$ & $74.81 \pm 0.25$ \\

RCAM + GELU & $68.03 \pm 0.36$ & $70.15 \pm 0.10$ & $73.23 \pm 0.06$ & $73.41 \pm 0.35$ & $74.26 \pm 0.55$ & $74.32 \pm 0.40$ & $75.92 \pm 0.12$ & $76.31 \pm 0.07$ \\

MBC + GELU & $68.19 \pm 0.26$ & $70.92 \pm 0.14$ & $73.82 \pm 0.22$ & $74.14 \pm 0.15$ & $75.10 \pm 0.38$ & $75.13 \pm 0.38$ & $76.41 \pm 0.33$ & $76.62 \pm 0.19$ \\


CKD + GELU & $70.38 \pm 1.06$ & $71.74 \pm 0.47$ & $73.44 \pm 0.65$ & $73.76 \pm 0.60$ & $74.25 \pm 0.58$ & $74.51 \pm 0.57$ & $74.72 \pm 0.39$ & $74.98 \pm 0.37$ \\

\midrule

ADC-RANet & $\mathbf{70.69 \pm 0.24}$ & $\mathbf{72.59 \pm 0.21}$ & $\mathbf{74.91 \pm 0.11}$ & $\mathbf{75.09 \pm 0.12}$ & $\mathbf{76.24 \pm 0.07}$ & $\mathbf{76.23 \pm 0.28}$ & $\mathbf{76.93} \pm \mathbf{0.30}$ & $\mathbf{77.12} \pm \mathbf{0.12}$ \\



\bottomrule
\end{tabular}
\end{table*}

\begin{table}[!t]
\centering
\caption{ Top-1 accuracy (\%) across exits for different distillation teacher strategies on CIFAR-100 across three seeds.}
\label{tab:ckd_teacher}
\small
\setlength{\tabcolsep}{5pt}
\begin{tabular}{c|ccc}
\hline
\textbf{Exit} & \textbf{LAST} & \textbf{ALL} & \textbf{CKD} \\
\hline
@1 & $70.16 \pm 0.23$ & $70.40 \pm 0.49$ & $70.34 \pm 0.28$ \\
@2 & $72.16 \pm 0.26$ & $72.18 \pm 0.31$ & $72.07 \pm 0.02$ \\
@3 & $73.36 \pm 0.29$ & $73.30 \pm 0.36$ & $73.48 \pm 0.12$ \\
@4 & $73.47 \pm 0.14$ & $73.55 \pm 0.22$ & $73.56 \pm 0.13$ \\
@5 & $73.75 \pm 0.32$ & $73.99 \pm 0.41$ & $74.11 \pm 0.08$ \\
@6 & $73.72 \pm 0.23$ & $73.97 \pm 0.49$ & $74.25 \pm 0.16$ \\
@7 & $73.85 \pm 0.28$ & $74.30 \pm 0.44$ & $74.62 \pm 0.14$ \\
@8 & $74.02 \pm 0.19$ & $74.43 \pm 0.42$ & $74.81 \pm 0.25$ \\
\hline
\end{tabular}
\end{table}
\subsection{Main Results}
For CIFAR-100, the anytime results in Figure \ref{fig:cifar100anytime} show that the proposed architecture consistently outperforms RANet, MSDNet, and ensemble baselines across all exits and computational budgets. At approximately 18M FLOPs, our model reaches 70.69\% accuracy, a gain of 3.29\% over RANet, and at around 60M FLOPs, the intermediate exits reach 76.24\%, improving the original RANet by 3.12\%. The final exit attains 77.12\% accuracy at 98M FLOPs, which is 2.83\% higher than competing methods.

The budgeted batch results on ImageNet in Figure \ref{fig:imagenet-bb} exhibit similar trends, with the largest relative gains at low budgets. For a budget of approximately 766M FLOPs, our architecture attains 67.0\% accuracy, around 3\% above the original RANet, and preserves a better cost–effectiveness profile across the full range of computational budgets.

\subsection{Ablation Study}
In this section, we evaluate the contribution of each proposed modification, as well as their synergy, to the overall improvements over the original RANet.
Table \ref{tab:anytime_acc} summarizes the anytime prediction results for the different ablation settings. While individual components do not achieve the highest accuracy when used in isolation, their combination is shown to consistently improve performance, leading to better results on CIFAR-100 and ImageNet. All ablation experiments used the same setup previously defined.


\paragraph{Residual Channel Attention Mechanism.}
The introduction of the RCAM yields only a modest improvement in the accuracy of the initial exits, but its impact becomes more pronounced at the final exits of the network. For example, exits 7 and 8 achieve gains of 1.31\% and 1.40\%, respectively. 
Furthermore, combining RCAM with GELU yields an additional performance improvement. 
This larger improvement at the final exits is expected, as the channel-attention mechanism filters and weights the most relevant channels as the subnetwork resolution increases throughout the network. Consequently, high-resolution subnetworks benefit more from the channel-attention mechanism.


To further verify the importance of feature reuse in RANet, we conducted an additional experiment removing the ConvUp operation, so that each subnetwork relies solely on its own features; this degrades performance substantially, with later exits affected the most (see supplementary material). We also compared RCAM applied before versus after feature fusion, confirming that incorporating context from both the current and previous subnetwork is the more effective fusion strategy (see supplementary material).


\paragraph{GELU.}


We evaluated the full CIFAR-100 test set at the channel level and found that ReLU leaves $3.33\% \pm 1.40\%$ of neurons permanently dead, compared with only $0.32\% \pm 0.24\%$ under GELU ($n=3$ seeds). All dead neurons were found in later ConvUp operations. This reduction is consistent with GELU's higher top-1 accuracy in Table~\ref{tab:anytime_acc}, suggesting that ReLU may limit feature reuse in the later stages of RANet. The modifications are complementary, addressing different stages of the exit sequence: knowledge distillation primarily improves early exits, while RCAM, GELU, and MixUp contribute more strongly to later-exit performance.

\paragraph{Cascaded Knowledge Distillation.}

Adding CKD to the original RANet improves accuracy across all exits and subnetworks, as shown in Table \ref{tab:anytime_acc}. In particular, the first exit gains approximately 3.0 percentage points from the distillation loss alone, with the largest improvements generally occurring at earlier exits. These gains come at no additional inference cost, since distillation only affects training. This suggests that CKD helps early-exit classifiers learn more discriminative decision boundaries.

To further investigate the choice of teacher, Table \ref{tab:ckd_teacher} compares three distillation configurations: \textit{last}, which uses only the final exit as the teacher; \textit{all}, which uses all subsequent classifiers; and \textit{CKD}, which relies on neighboring classifiers. While the three approaches achieve comparable performance at early exits, CKD yields more consistent improvements at later exits. It also exhibits lower variability across runs, as indicated by its standard deviations, suggesting more stable training.





\paragraph{MixUp-Based Calibration.} 

As shown in Tables \ref{tab:signed_bias_comparison} and \ref{tab:anytime_acc}, MixUp improves performance across all exits, with the largest gains at later classifiers. It also improves calibration by reducing signed bias and overconfidence, leading to more reliable exit decisions. Notably, exits 7 and 8 achieve accuracy gains of approximately 1.92\% and 1.96\%, respectively, over the baseline final exit.

The ablation study in Table \ref{tab:exit_usage_accuracy_mixup} further shows that MixUp improves accuracy at every exit. By reducing overconfidence at early exits, fewer samples are prematurely assigned to exit 1, allowing more to reach subsequent classifiers and resulting in better accuracy and a more balanced distribution of samples across exits.


\begin{table}[t]
\centering
\caption{Exit acceptance and accuracy across exits for Baseline and MixUp for a fixed 80\% confidence threshold across 3 seeds.}
\label{tab:exit_usage_accuracy_mixup}
\small
\setlength{\tabcolsep}{3pt}
\begin{tabular}{c|cc|cc}
\hline
& \multicolumn{2}{c|}{\textbf{Baseline (RANet)}} & \multicolumn{2}{c}{\textbf{RANet + MixUp}} \\
\textbf{Exit} & \textbf{Sample (\%)} & \textbf{Accuracy (\%)} & \textbf{Sample (\%)} & \textbf{Accuracy (\%)} \\
\hline
@1 & $57.97\!\pm\!0.42$ & $87.73\!\pm\!0.42$ & $41.33\!\pm\!2.99$ & $95.10\!\pm\!0.92$ \\
@2 & $17.07\!\pm\!0.12$ & $65.63\!\pm\!1.39$ & $12.93\!\pm\!1.00$ & $85.67\!\pm\!3.66$ \\
@3 & $9.37\!\pm\!0.31$ & $59.07\!\pm\!2.01$ & $9.00\!\pm\!0.85$ & $80.47\!\pm\!3.71$ \\
@4 & $4.17\!\pm\!0.21$ & $47.20\!\pm\!0.87$ & $4.27\!\pm\!0.25$ & $72.13\!\pm\!3.04$ \\
@5 & $3.00\!\pm\!0.17$ & $45.73\!\pm\!3.22$ & $3.33\!\pm\!0.50$ & $69.80\!\pm\!5.71$ \\
@6 & $1.67\!\pm\!0.06$ & $43.90\!\pm\!1.95$ & $1.90\!\pm\!0.00$ & $65.33\!\pm\!5.00$ \\
@7 & $1.43\!\pm\!0.12$ & $44.83\!\pm\!3.27$ & $2.43\!\pm\!0.15$ & $72.60\!\pm\!1.37$ \\
@8 & $5.30\!\pm\!0.10$ & $26.70\!\pm\!0.36$ & $24.80\!\pm\!5.12$ & $42.20\!\pm\!3.03$ \\
\hline
\end{tabular}
\end{table}

\begin{table}[t]
\centering
\caption{ Anytime Top-1 accuracy (\%) across exits for ADC-RANet and RANet on ImageNet.}
\label{tab:flops_matched_acc}
\small
\setlength{\tabcolsep}{5pt}
\begin{tabular}{c|cc|cc}
\hline
\multirow{2}{*}{\textbf{Exit}} & \multicolumn{2}{c|}{\textbf{FLOPs (M)}} & \multicolumn{2}{c}{\textbf{Top-1 Acc. (\%)}} \\
 & \textbf{ADC-RANet} & \textbf{RANet} & \textbf{ADC-RANet} & \textbf{RANet} \\
\hline
@1 & 550.78  & 550.78  & 66.096 & 64.603 \\
@2 & 1004.48 & 1004.48 & 70.108 & 68.385 \\
@3 & 1270.38 & 1269.16 & 72.344 & 70.159 \\
@4 & 1469.47 & 1468.26 & 73.280 & 70.873 \\
@5 & 2401.69 & 2395.72 & 75.840 & 74.093 \\
@6 & 2710.25 & 2704.28 & 76.184 & 74.323 \\
@7 & 3361.27 & 3351.32 & 75.892 & 74.451 \\
@8 & 3459.24 & 3449.29 & 76.088 & 74.755 \\
\hline
\end{tabular}
\end{table}

Finally, our method also transfers well to ImageNet, as shown in Table \ref{tab:flops_matched_acc}, which reports the anytime performance results. We observe a favorable trade-off between computational cost and accuracy gains, with our lightweight component, discussed in the previous section, introducing only a marginal computational overhead while improving accuracy.

\section{Gradient Conflict and Loss Function}

We further investigate the interactions between early and late exits and their impact on performance. 
 Since the gradients of high- and low-resolution classifiers are poorly aligned (low cosine similarity) in the default RANet training (more details in the supplementary material), we investigate how this misalignment drives performance trade-offs.
We analyze gradient conflicts by modifying the loss: we apply cross-entropy only to selected classifiers while using knowledge distillation for all. This allows us to control each classifier’s contribution and exposes a clear trade-off: prioritizing cross-entropy on early, low-resolution classifiers boosts their accuracy but hurts later classifiers, and vice versa, revealing competing objectives across network depths.

\subsection{Gradient Conflict on RANet}





The original loss function of RANet is defined as the sum of cross-entropy losses over all classifiers \cite{yang2020resolution}. 
However, jointly training all exits can cause optimization conflicts, since classifiers at different depths and scales operate at different input resolutions and semantic levels and therefore benefit from distinct feature representations.
To demonstrate this, we conducted experiments showing that in RANet models, the addition of high-resolution gradients degrades the performance of low-resolution ones. 


We trained four RANet variants on CIFAR-100, each consisting of four subnetworks processing resolutions of 8×8, 16×16, 32×32, and 32×32. The first model trains only the 8×8 classifiers, the second adds 16×16 classifiers, and so on. Additionally, we maintained the same first-layer resolution reduction across all networks to ensure identical pooling operations in every experiment.

\begin{figure}[t]
  \centering
   \includegraphics[width=1.0\linewidth]{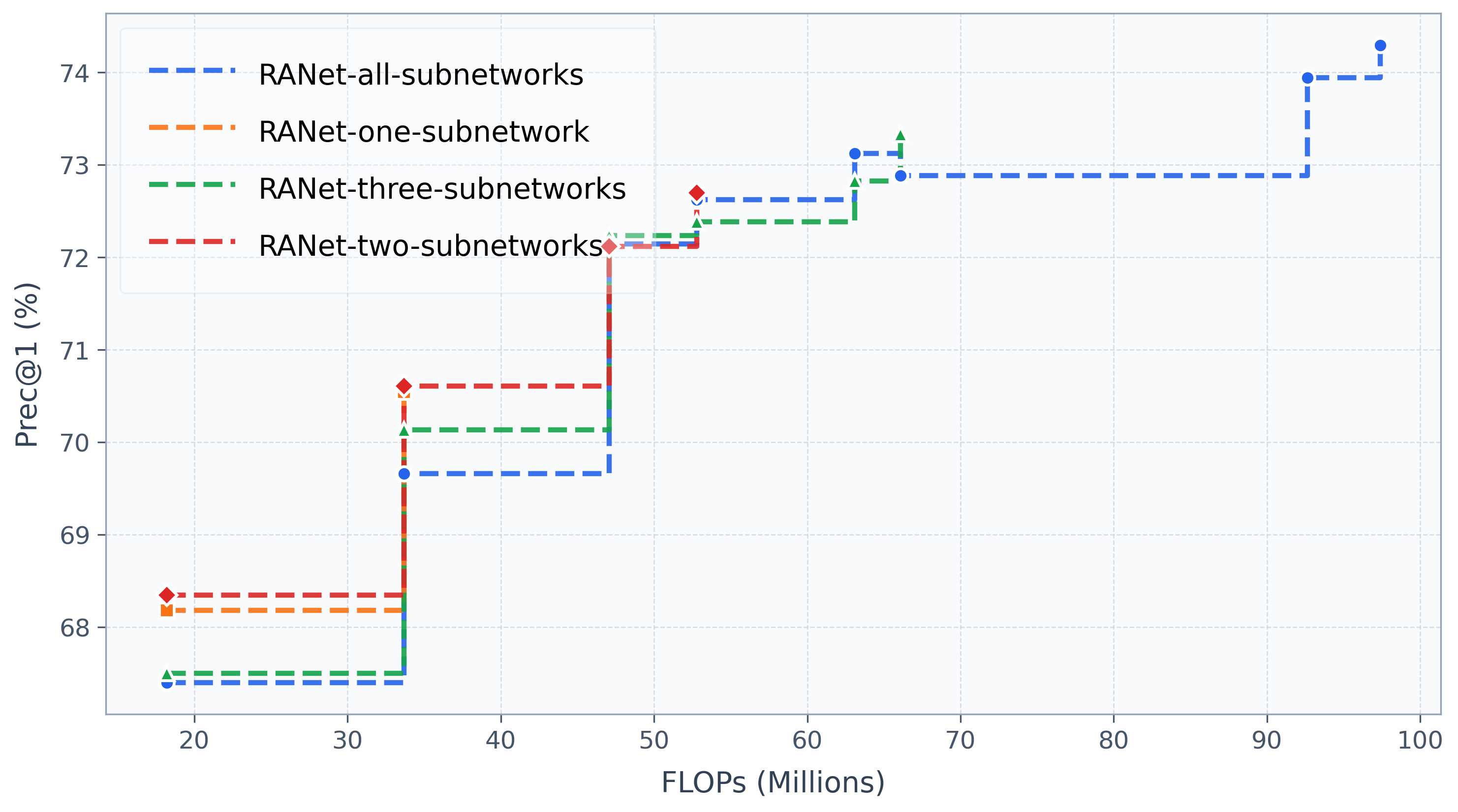}
    
   \caption{Gradient conflict when training different subnetwork configurations. The different RANets have a different number of exits, and as the number of exits with high resolution increases, the performance for early exits drops}
   \label{fig:conflict}
\end{figure}

Figure \ref{fig:conflict} illustrates the gradient conflict across scales. For the network trained exclusively with the low-resolution subnetwork, the classifier at approximately 18 million FLOPs achieves 68.3\% accuracy, whereas the same subnetwork trained with all classifiers across different resolutions achieves 67.5\%. Similarly, at the second exit with a budget of approximately 35 million FLOPs, the low-resolution-only network achieves 70.7\%, while the network trained with all resolutions achieves 69.7\%. These results demonstrate that introducing high-resolution classifiers can reduce the performance of initial exits by up to 1\%.


We adapted DFS \cite{gong2024deep} to RANet to reduce gradient conflict by separating features into exit-specific and shared groups. DFS improves the initial exit by 1.20\% (67.40\% to 68.60\%) but degrades the final exit by 0.20\% (74.29\% to 74.09\%). We attribute this behavior to the difference in feature reuse between MSDNet and RANet: in RANet's coarse-to-fine architecture, partitioning features at an early exit prevents later exits from updating the exit-specific representations. Moreover, combining DFS with CKD and MixUp underperforms CKD+MixUp alone by up to 1.07\% at the final exit (see supplementary material).

\subsection{Classifier Trade-off}
\label{sec:trade_off}
We analyze how classifier performance across different depths changes when the loss function is manipulated. Specifically, we apply cross-entropy loss only to a subset of classifiers while using knowledge distillation for all of them, which reveals a trade-off: increasing the cross-entropy weight on early low-resolution classifiers degrades the performance of later classifiers, and vice versa.




To show this trade-off, four experiments are conducted: one applying cross-entropy loss only to the highest-resolution subnetworks, one using equal cross-entropy weighting across all classifiers, and finally, assigning cross-entropy weighting to the two lowest-resolution classifiers. For all experiments, we use the ADC-RANet defined in Section \ref{overview}, by changing only the loss function. The adapted loss function multiplies each cross-entropy term by a weight $w_n$.
\begin{equation}
\label{eq:weighted_loss}
\mathcal{L}(x,y)
=
\sum_{n=1}^{N}
w_n \, \mathcal{L}_{\mathrm{CE}}(p_n,y)
+
\frac{\beta T^2}{N - 1}
\sum_{k=1}^{N-1}
\mathrm{KL}
\!\left(
p_{k+1}^{T}
\,\|\,
p_k^{T}
\right),
\end{equation}



\begin{figure}[t]
\label{fig3}
  \centering
   \includegraphics[width=1.0\linewidth]{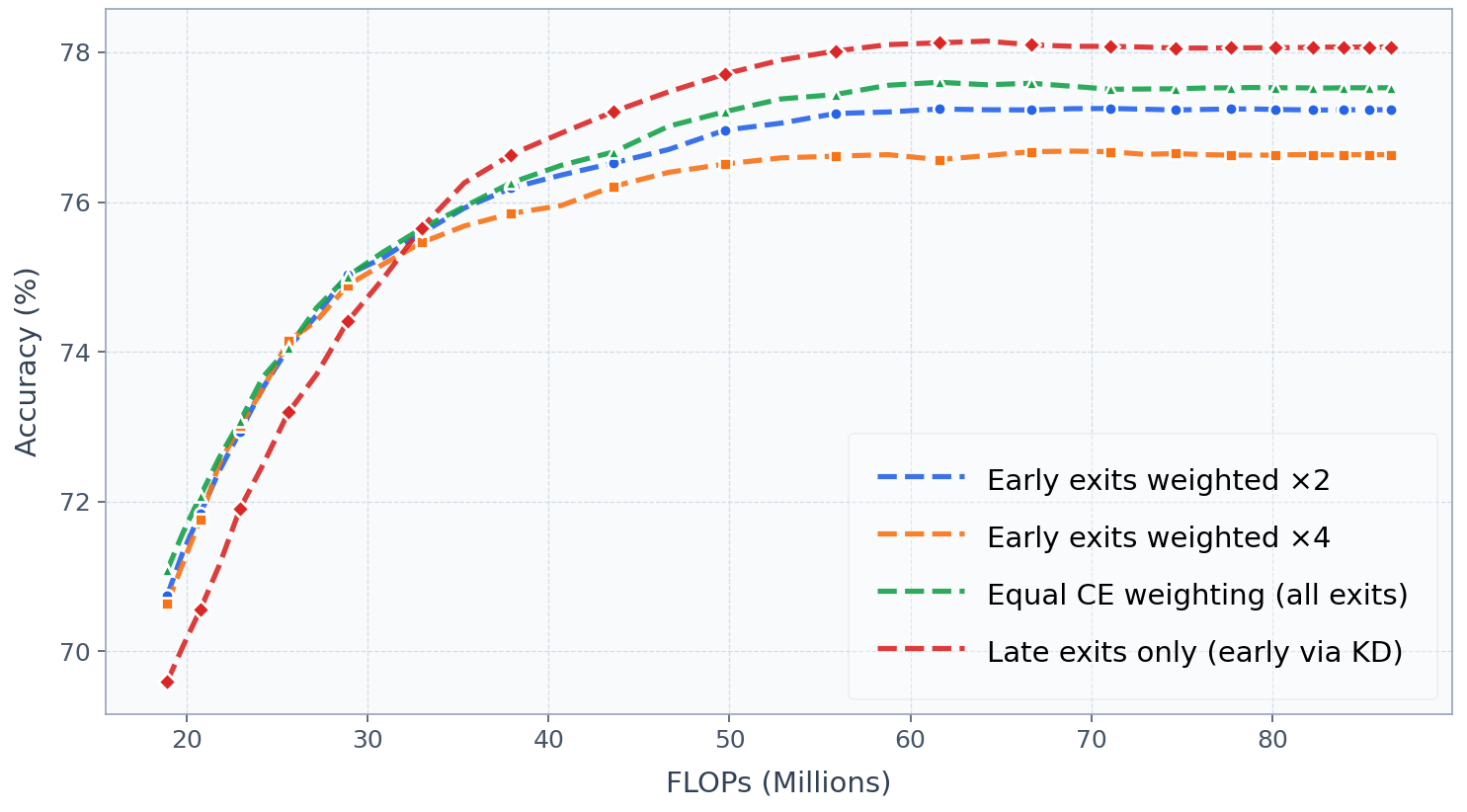}
    
   \caption{Budgeted batch for different loss training configurations under different computational budgets}
   \label{fig:onecol}
\end{figure}

Figure~\ref {fig:onecol} shows the budgeted batch curves for the four experiments. When only late exits are trained with cross-entropy and early exits rely on distillation, the early classifiers perform slightly worse, while the final classifiers achieve the best performance, reaching approximately 78\% at 60M FLOPs compared with 77\% for the default ADC-RANet with equal loss weighting. Adding cross-entropy to the early exits improves their performance but reduces that of the final classifiers.
Increasing the weight of early-exit losses to $w_n=2$ and $4$ further improves neither side substantially: early-exit performance largely plateaus, while late-exit performance decreases more noticeably. This asymmetric trade-off likely results from architectural limitations related to input resolution and network depth.


We also show that all early classifiers can be trained solely through cascaded distillation. Although these classifiers perform slightly worse than those trained with both cross-entropy and distillation, this strategy yields stronger late classifiers.

Overall, the optimal loss configuration depends on the deployment objective. Weighting early exits is preferable under strict computational constraints, while distillation-only training favors late-exit accuracy. Equal weighting provides the best compromise between early and late exits.

\section{Conclusion}


In this work, we improve multi-exit neural networks on resource-constrained devices through cost-efficient enhancements to RANet. We propose RANet with Attention, Distillation and Calibration (ADC-RANet), which combines a lightweight cross-scale channel-attention mechanism, a top-down cascaded knowledge distillation strategy, and a training framework combined with MixUp regularization that reduces overconfidence. Together, these components improve early-exit performance and cross-scale information sharing while preserving the efficiency that makes multi-exit models suitable for embedded inference. Unlike heavier mechanisms such as cross-attention, our method maintains low cost while improving accuracy at both early and late exits. RCAM strengthens bottom-up relationships between scales, CKD introduces top-down knowledge transfer, enabling early classifiers to learn from late exits, and MixUp further improves classifier calibration and generalization.
%

 ADC-RANet yields gains of 3.3\% and 2.8\% on CIFAR-100 for the first and last classifiers, respectively. It also outperforms MSDNet in the budgeted batch setting: for 1000M FLOPs it improves accuracy by 3.4\%, and for 2000M FLOPs by 1.63\% on ImageNet.

\noindent\textbf{Supplementary materials.}
Includes details on how GELU improves ADC-RANet’s accuracy by reducing dead neurons in ConvUp operations, possible techniques for reducing gradient conflict in RANet, the design of RCAM, and reliability diagrams showing how ADC-RANet is less overconfident.

\paragraph{Acknowledgements} This work was supported by Distech Controls Inc., the Natural Sciences and Engineering Research Council of Canada, the Digital Research Alliance of Canada, and MITACS.

{
    \small
    \bibliographystyle{ieeenat_fullname}
    \bibliography{main}
}

\clearpage
\appendix
\maketitlesupplementary
\section{GELU in ConvUp operations}

We propose replacing RANet’s default ReLU activation function \cite{nairhinton2010relu} with GELU \cite{gelu} to mitigate dead neurons \cite{dyingrelu} in ConvUp operations and leverage GELU’s smoothness around zero, enabling more effective feature exploitation across scales within subnetworks.

We evaluated the complete CIFAR-100 test set on RANets at the channel level using ReLU and GELU to demonstrate the impact of replacing the default ReLU activation function with GELU. We define a permanently dead neuron as one that had no absolute activation value greater than $10^{-3}$ in all samples. Under the ReLU activation function, $3.33\% \pm 1.40\%$ ($n=3$ seeds) of neurons were permanently dead, and under GELU, $0.32\% \pm 0.24\%$ ($n=3$ seeds) were permanently dead. All dead neurons were ConvUp in later layers, suggesting that ReLU can impair feature reuse in the later stages of RANet by completely suppressing a subset of channels. In addition, the smoother nonlinear transformation provided by GELU may further benefit optimization and feature representation, as supported by its performance over Leaky ReLU at all but the two earliest exits, where the two activations are comparable.

\begin{table}[t]
\centering
\small
\setlength{\tabcolsep}{4pt}
\caption{Top-1 accuracy (\%) across exits for different configurations for three runs.}
\label{tab:prec1}
\begin{tabular}{lccc}
\toprule
Exit & ReLu & GELU & Leaky Relu \\
\midrule
1 & $67.40 \pm 0.31$ & $67.33 \pm 0.42$ & $67.60 \pm 0.21$ \\
2 & $69.66 \pm 0.37$ & $69.82 \pm 0.34$ & $70.02 \pm 0.32$ \\
3 & $72.14 \pm 0.27$ & $72.70 \pm 0.30$ & $72.28 \pm 0.35$ \\
4 & $72.62 \pm 0.36$ & $73.14 \pm 0.67$ & $72.55 \pm 0.27$ \\
5 & $73.12 \pm 0.44$ & $73.82 \pm 0.49$ & $73.31 \pm 0.34$ \\
6 & $72.88 \pm 0.18$ & $73.84 \pm 0.25$ & $73.47 \pm 0.45$ \\
7 & $73.94 \pm 0.42$ & $74.76 \pm 0.30$ & $74.80 \pm 0.43$ \\
8 & $74.29 \pm 0.13$ & $75.41 \pm 0.11$ & $74.91 \pm 0.30$ \\
\bottomrule
\end{tabular}
\end{table}
\section{Gradient Conflict}

\cite{gong2024deep} proposes a strategy for reducing the gradient conflict in multi-exit networks by partitioning the features into two groups: one group is exit-specific and is updated only by the loss of its exit, and the other is shared and updated by the losses of multiple classifiers; in the forward pass, all features are used. The goal of this strategy is to allow exit-specific features to be optimized for their corresponding classifiers, while shared features continue to benefit from information aggregated across all exits.

We empirically adapted DFS to RANet, as it had not previously been tested in this architecture; the results are shown in Table \ref{tab:dfs_comparison}. DFS improves accuracy at the first two exits but degrades it at every exit thereafter. We attribute this reversal to a structural difference in how MSDNet and RANet reuse features across exits. In MSDNet, the DFS is applied to the coarse features where the classifiers are located, while the fine-to-coarse features remain unchanged; in RANet, because of the coarse-to-fine architecture, partitioning the features at an early exit prevents the loss of later exits from updating the exit-specific portion of those representations. Notably, combining DFS with CKD and MixUp still underperforms CKD+MixUp alone on most exits, indicating how DFS hurts performance with the additional components.

 In addition, we compute and compare gradient metrics for different configurations of RANet. Table \ref{tab:grad_conflict} shows that the initial RANet exit is not well aligned in direction with subsequent exits, and its gradient norm is the largest. Also, the weighting strategy does not eliminate the conflict, but moves it. Table \ref{tab:grad_conflict_kd} indicates how the CKD improves the cosine similarity with few negative conflicting batches.  

 \begin{table}[t]
\centering
\caption{ Gradient conflict between the first exit and later exits on CIFAR-100 for RANet, RANet with CKD, and ADC-RANet for different losses. We show the cosine similarity between the cross-entropy gradients of exit 1 and exit $k$ on their shared parameters, the fraction of mini-batches with negative cosine similarity, and the gradient norms.}
\label{tab:grad_conflict}
\setlength{\tabcolsep}{4pt}
\begin{tabular}{lccccccc}
\toprule
Exit $k$ & 2 & 3 & 4 & 5 & 6 & 7 & 8 \\
\midrule
\multicolumn{8}{l}{\textit{RANet}} \\
$\cos(g_1, g_k)$ & 0.24 & 0.15 & 0.12 & 0.10 & 0.12 & 0.10 & 0.09 \\
Neg. frac. & 0.06 & 0.17 & 0.20 & 0.23 & 0.19 & 0.21 & 0.27 \\
$\|g_1\|$ & 1.99 & 1.99 & 1.99 & 1.99 & 1.99 & 1.99 & 1.99 \\
$\|g_k\|$ & 1.07 & 0.58 & 0.47 & 0.38 & 0.34 & 0.17 & 0.16 \\
\midrule
\multicolumn{8}{l}{\textit{RANet + CKD}} \\
$\cos(g_1, g_k)$ & 0.43 & 0.24 & 0.23 & 0.19 & 0.19 & 0.14 & 0.14 \\
Neg. frac. & 0.00 & 0.01 & 0.03 & 0.00 & 0.00 & 0.04 & 0.07 \\
$\|g_1\|$ & 2.08 & 2.08 & 2.08 & 2.08 & 2.08 & 2.08 & 2.08 \\
$\|g_k\|$ & 1.36 & 0.85 & 0.77 & 0.67 & 0.65 & 0.42 & 0.40 \\
\midrule
\multicolumn{8}{l}{\textit{ADC-RANet, late exits only}} \\
$\cos(g_1, g_k)$ & 0.16 & 0.10 & 0.10 & 0.03 & 0.03 & 0.05 & 0.04 \\
Neg. frac. & 0.09 & 0.18 & 0.17 & 0.35 & 0.38 & 0.24 & 0.24 \\
$\|g_1\|$ & 0.06 & 0.06 & 0.06 & 0.06 & 0.06 & 0.06 & 0.06 \\
$\|g_k\|$ & 0.72 & 0.06 & 0.32 & 0.27 & 0.15 & 0.06 & 0.06 \\
\midrule
\multicolumn{8}{l}{\textit{ADC-RANet, equal weighting}} \\
$\cos(g_1, g_k)$ & 0.61 & 0.32 & 0.34 & 0.21 & 0.21 & 0.07 & 0.04 \\
Neg. frac. & 0.00 & 0.00 & 0.00 & 0.03 & 0.01 & 0.27 & 0.35 \\
$\|g_1\|$ & 0.70 & 0.70 & 0.70 & 0.70 & 0.70 & 0.70 & 0.70 \\
$\|g_k\|$ & 0.89 & 0.21 & 0.58 & 0.26 & 0.24 & 0.06 & 0.06 \\
\midrule
\multicolumn{8}{l}{\textit{ADC-RANet, early exits weighted $\times2$}} \\
$\cos(g_1, g_k)$ & 0.78 & 0.45 & 0.43 & 0.22 & 0.23 & 0.15 & 0.16 \\
Neg. frac. & 0.00 & 0.00 & 0.00 & 0.02 & 0.01 & 0.07 & 0.03 \\
$\|g_1\|$ & 2.20 & 2.20 & 2.20 & 2.20 & 2.20 & 2.20 & 2.20 \\
$\|g_k\|$ & 2.28 & 1.25 & 1.28 & 0.22 & 0.19 & 0.04 & 0.06 \\
\midrule
\multicolumn{8}{l}{\textit{ADC-RANet, early exits weighted $\times4$}} \\
$\cos(g_1, g_k)$ & 0.75 & 0.53 & 0.55 & 0.43 & 0.35 & 0.11 & 0.12 \\
Neg. frac. & 0.00 & 0.00 & 0.00 & 0.00 & 0.00 & 0.21 & 0.21 \\
$\|g_1\|$ & 1.98 & 1.98 & 1.98 & 1.98 & 1.98 & 1.98 & 1.98 \\
$\|g_k\|$ & 2.09 & 1.52 & 1.63 & 0.59 & 0.43 & 0.11 & 0.10 \\
\bottomrule
\end{tabular}
\end{table}

\begin{table}[t]
\centering
\caption{Gradient conflict between the first exit and later exits on CIFAR-100 for RANet and for the distillation teacher strategies. We show the cosine similarity between the cross-entropy gradients of exit 1 and exit $k$ on their shared parameters, the fraction of mini-batches with negative cosine similarity, and the gradient norms.}
\label{tab:grad_conflict_kd}
\setlength{\tabcolsep}{4pt}
\begin{tabular}{lccccccc}
\toprule
Exit $k$ & 2 & 3 & 4 & 5 & 6 & 7 & 8 \\
\midrule
\multicolumn{8}{l}{\textit{RANet}} \\
$\cos(g_1, g_k)$ & 0.24 & 0.15 & 0.12 & 0.10 & 0.12 & 0.10 & 0.09 \\
Neg. frac. & 0.06 & 0.17 & 0.20 & 0.23 & 0.19 & 0.21 & 0.27 \\
$\|g_1\|$ & 1.99 & 1.99 & 1.99 & 1.99 & 1.99 & 1.99 & 1.99 \\
$\|g_k\|$ & 1.07 & 0.58 & 0.47 & 0.38 & 0.34 & 0.17 & 0.16 \\
\midrule
\multicolumn{8}{l}{\textit{RANet + LAST}} \\
$\cos(g_1, g_k)$ & 0.30 & 0.17 & 0.16 & 0.15 & 0.14 & 0.14 & 0.12 \\
Neg. frac. & 0.02 & 0.12 & 0.12 & 0.14 & 0.15 & 0.10 & 0.19 \\
$\|g_1\|$ & 2.05 & 2.05 & 2.05 & 2.05 & 2.05 & 2.05 & 2.05 \\
$\|g_k\|$ & 1.31 & 0.80 & 0.67 & 0.53 & 0.52 & 0.31 & 0.31 \\
\midrule
\multicolumn{8}{l}{\textit{RANet + ALL}} \\
$\cos(g_1, g_k)$ & 0.40 & 0.26 & 0.24 & 0.20 & 0.18 & 0.14 & 0.14 \\
Neg. frac. & 0.00 & 0.03 & 0.03 & 0.07 & 0.07 & 0.12 & 0.16 \\
$\|g_1\|$ & 2.12 & 2.12 & 2.12 & 2.12 & 2.12 & 2.12 & 2.12 \\
$\|g_k\|$ & 1.46 & 0.90 & 0.81 & 0.72 & 0.71 & 0.48 & 0.47 \\
\midrule
\multicolumn{8}{l}{\textit{RANet + CKD}} \\
$\cos(g_1, g_k)$ & 0.43 & 0.24 & 0.23 & 0.19 & 0.19 & 0.14 & 0.14 \\
Neg. frac. & 0.00 & 0.01 & 0.03 & 0.00 & 0.00 & 0.04 & 0.07 \\
$\|g_1\|$ & 2.08 & 2.08 & 2.08 & 2.08 & 2.08 & 2.08 & 2.08 \\
$\|g_k\|$ & 1.36 & 0.85 & 0.77 & 0.67 & 0.65 & 0.42 & 0.40 \\
\bottomrule
\end{tabular}
\end{table}

\begin{table}[t]
\centering
\caption{Top-1 accuracy across exits for RANet, DFS+RANet, CKD+MixUp, and DFS+CKD+MixUp.}
\label{tab:dfs_comparison}
\small
\setlength{\tabcolsep}{1pt}
\begin{tabular}{c|cccc}
\hline
\textbf{Exit} & \textbf{RANet} & \textbf{DFS+RANet} & \textbf{CKD+MixUp} & \textbf{DFS+CKD+MixUp} \\
\hline
@1 & $67.40 \pm 0.31$ & $68.60 \pm 0.26$ & $69.95 \pm 0.05$ & $70.48 \pm 0.35$ \\
@2 & $69.66 \pm 0.37$ & $70.06 \pm 0.31$ & $72.45 \pm 0.14$ & $71.71 \pm 0.34$ \\
@3 & $72.14 \pm 0.27$ & $71.58 \pm 0.58$ & $74.28 \pm 0.22$ & $73.10 \pm 0.41$ \\
@4 & $72.62 \pm 0.36$ & $72.01 \pm 0.36$ & $74.54 \pm 0.37$ & $73.23 \pm 0.46$ \\
@5 & $73.12 \pm 0.44$ & $72.19 \pm 0.33$ & $75.42 \pm 0.20$ & $73.76 \pm 0.51$ \\
@6 & $72.88 \pm 0.18$ & $72.77 \pm 0.50$ & $75.42 \pm 0.15$ & $73.81 \pm 0.37$ \\
@7 & $73.94 \pm 0.42$ & $73.69 \pm 0.57$ & $76.20 \pm 0.27$ & $75.14 \pm 0.25$ \\
@8 & $74.29 \pm 0.13$ & $74.09 \pm 0.55$ & $76.32 \pm 0.25$ & $75.25 \pm 0.27$ \\
\hline
\end{tabular}
\end{table}


\section{RCAM}

\cite{dai2021attentional} shows that the use of simple concatenation operations for feature fusion may not be the best option to fuse features of different scales; in the RANet context, the fusion operation is a simple concatenation followed by channel compression, leading to a context-unaware fusion where the lower-resolution and high-resolution features are treated equally. To address this, we introduce RCAM in a fully context-aware setup, where it receives the concatenation of the low- and high-resolution features. 

Table \ref{tab:convup} shows the RANet results when removing the ConvUp operations. The accuracies are not monotonically increasing across the exits. This table further shows that the later classifiers benefit more from feature reuse, highlighting the importance of shared representations for their performance.

In addition, following a similar setup from \cite{dai2021attentional}, we place the RCAM in two different places: (i) applied after the fusion operation, where the RCAM sees features of different resolutions, and (ii) applied only on the features from the previous operations, where the RCAM doesn't have the full context of the resolutions. The results are shown in Table \ref{tab:rcam_comparison}; We observe that applying RCAM after fusion yields the best overall performance, indicating that access to the full feature context across different subnetworks is more effective for feature reuse in RANet. Applying RCAM before fusion provides a slight advantage at Exit~4.

Table \ref{tab:flops_matched_acc} indicates the FLOP difference between ADC-RANet and RANet configuration for CIFAR-100 dataset.

\begin{table}[t]
\centering
\caption{Top-1 accuracy (\%) across exits on CIFAR-100 across three seeds for RANet without feature reuse.}
\label{tab:convup}
\small
\setlength{\tabcolsep}{5pt}
\begin{tabular}{c|c}
\hline
\textbf{Exit} & \textbf{Accuracy} \\
\hline
@1 & $68.32 \pm 0.30$ \\
@2 & $69.87 \pm 0.33$ \\
@3 & $69.11 \pm 0.13$ \\
@4 & $69.82 \pm 0.25$ \\
@5 & $67.23 \pm 0.23$ \\
@6 & $67.85 \pm 0.40$ \\
@7 & $68.94 \pm 0.41$ \\
@8 & $69.93 \pm 0.47$ \\
\hline
\end{tabular}
\end{table}

\begin{table}[t]
\centering
\caption{Top-1 accuracy (\%) across exits for RCAM (post-fusion), RCAM (pre-fusion), and Baseline (RANet) on CIFAR-100 across three seeds.}
\label{tab:rcam_comparison}
\small
\setlength{\tabcolsep}{3pt}
\begin{tabular}{c|ccc}
\hline
\textbf{Exit} & \textbf{RCAM} & \textbf{RCAM (pre-fusion)} & \textbf{Baseline (RANet)} \\
\hline
@1 & $67.66\pm0.03$ & $66.87\pm0.41$ & $67.40\pm0.31$ \\
@2 & $70.31\pm0.46$ & $69.66\pm0.66$ & $69.66\pm0.37$ \\
@3 & $72.56\pm0.23$ & $72.48\pm0.28$ & $72.14\pm0.27$ \\
@4 & $72.73\pm0.06$ & $72.92\pm0.42$ & $72.62\pm0.36$ \\
@5 & $73.68\pm0.32$ & $73.63\pm0.40$ & $73.12\pm0.44$ \\
@6 & $73.82\pm0.16$ & $73.86\pm0.34$ & $72.88\pm0.18$ \\
@7 & $75.25\pm0.17$ & $74.88\pm0.39$ & $73.94\pm0.42$ \\
@8 & $75.69\pm0.10$ & $75.33\pm0.23$ & $74.29\pm0.13$ \\
\hline
\end{tabular}
\end{table}

\begin{table}[t]
\centering
\caption{ Anytime Top-1 accuracy (\%) across exits for ADC-RANet and RANet on CIFAR-100. Results are reported as mean $\pm$ standard deviation.}
\label{tab:flops_matched_acc}
\small
\setlength{\tabcolsep}{3pt}
\begin{tabular}{c|cc|cc}
\hline
\multirow{2}{*}{\textbf{Exit}}    & \multicolumn{2}{c|}{\textbf{FLOPs (M)}} & \multicolumn{2}{c}{\textbf{Top-1 Acc. (\%)}} \\
 & \textbf{ADC-RANet} & \textbf{RANet} & \textbf{ADC-RANet} & \textbf{RANet} \\
\hline
@1 & 18.20 & 18.20 & $\mathbf{70.69 \pm 0.24}$ & $67.40 \pm 0.31$ \\
@2 & 33.68 & 33.68 & $\mathbf{72.59 \pm 0.21}$ & $69.66 \pm 0.37$ \\
@3 & 47.12 & 47.07 & $\mathbf{74.91 \pm 0.11}$ & $72.14 \pm 0.27$ \\
@4 & 52.83 & 52.78 & $\mathbf{75.09 \pm 0.12}$ & $72.62 \pm 0.36$ \\
@5 & 63.24 & 63.09 & $\mathbf{76.24 \pm 0.07}$ & $73.12 \pm 0.44$ \\
@6 & 66.25 & 66.09 & $\mathbf{76.23 \pm 0.28}$ & $72.88 \pm 0.18$ \\
@7 & 93.11 & 92.63 & $\mathbf{76.93 \pm 0.30}$ & $73.94 \pm 0.42$ \\
@8 & 97.85 & 97.37 & $\mathbf{77.12 \pm 0.12}$ & $74.29 \pm 0.13$ \\
\hline
\end{tabular}
\end{table}

\section{MixUp Calibration}

We additionally plot reliability diagrams for Exits~1 and~8 under two configurations to visualize the gap between confidence and accuracy across different confidence levels, using 15 bins.

\begin{figure}[t]
  \centering
   \includegraphics[width=1.0\linewidth]{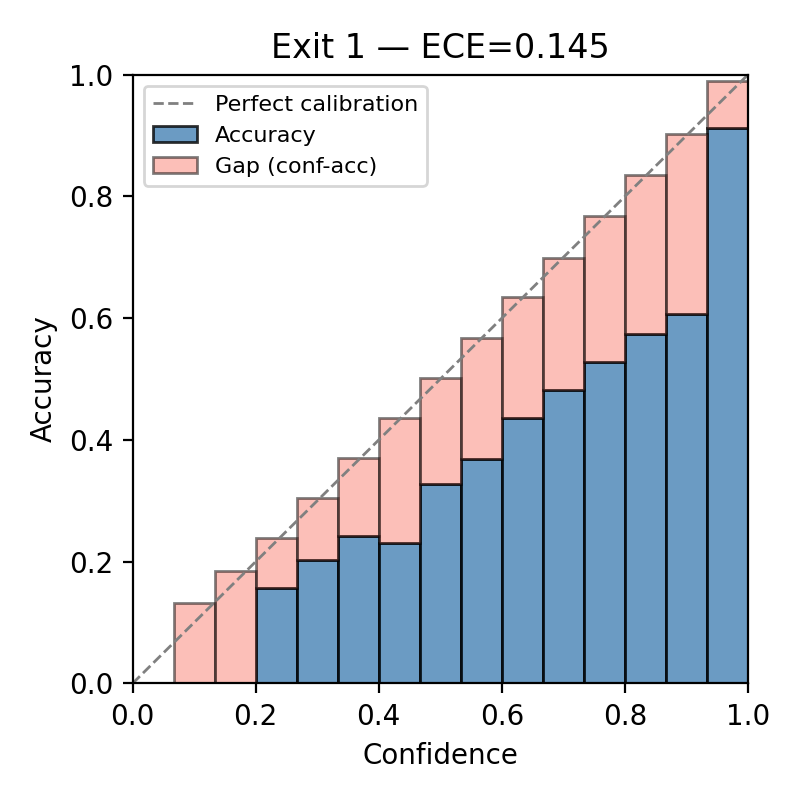}
    
   \caption{Reliability diagram for exit 1 of RANet}
   \label{fig:exit1}
\end{figure}

\begin{figure}[t]
  \centering
   \includegraphics[width=1.0\linewidth]{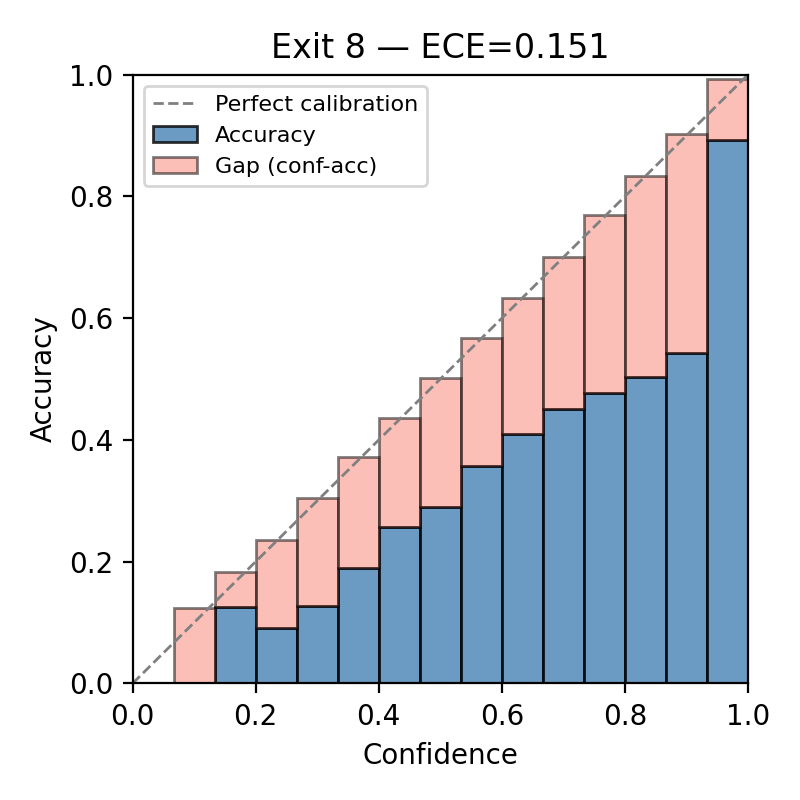}
    
   \caption{Reliability diagram for exit 8 of RANet}
   \label{fig:exit8}
\end{figure}

Figure \ref{fig:exit1} and Figure \ref{fig:exit8} show the gap between the accuracy and the expected accuracy for each bin, indicating overconfidence for those classifiers.

\begin{figure}[t]
  \centering
   \includegraphics[width=1.0\linewidth]{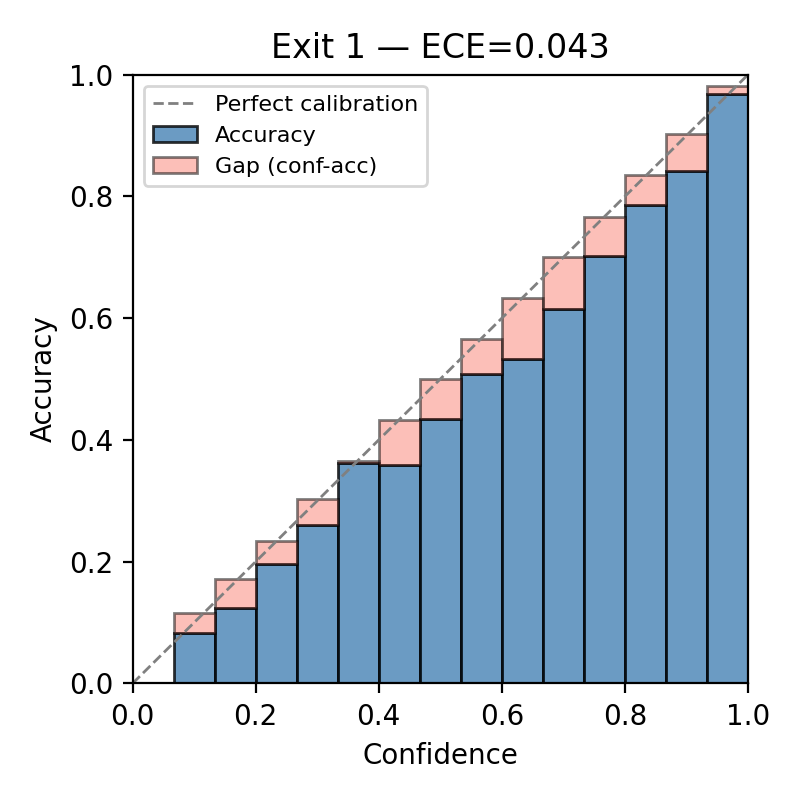}
    
   \caption{Reliability diagram for exit 1 of RANet + MixUp}
   \label{fig:exit1-mixup}
\end{figure}

\begin{figure}[t]
  \centering
   \includegraphics[width=1.0\linewidth]{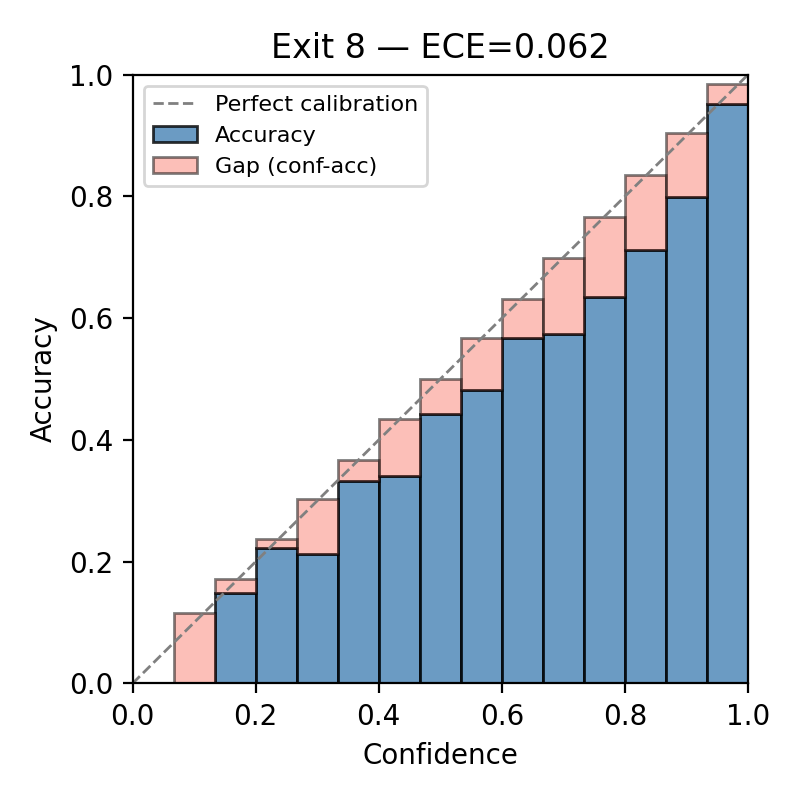}
    
   \caption{Reliability diagram for exit 8 of RANet + MixUp}
   \label{fig:exit8-mixup}
\end{figure}

Figures \ref{fig:exit1-mixup} and \ref{fig:exit8-mixup} show the gap between confidence and accuracy across bins with MixUp during training. MixUp substantially reduces this gap, resulting in more reliable classifiers.

\end{document}